\documentclass[lettersize,journal]{IEEEtran}
\usepackage{amsmath,amsfonts}
\usepackage{algorithmic}
\usepackage{algorithm}
\usepackage{array}
\usepackage[caption=false,font=normalsize,labelfont=sf,textfont=sf]{subfig}
\usepackage{textcomp}
\usepackage{stfloats}
\usepackage{url}
\usepackage{verbatim}
\usepackage{graphicx}
\usepackage{cite}

\begin{document}

\title{Semantic-Aware Cooperative Communication and Computation Framework in Vehicular Networks}

\author{Jingbo Zhang, Maoxin Ji,~\and Qiong Wu~\IEEEmembership{Senior Member,~IEEE,}~\and \\Pingyi Fan~\IEEEmembership{Senior Member,~IEEE,}, Kezhi Wang,~\IEEEmembership{Senior Member,~IEEE}, Wen Chen,~\IEEEmembership{Senior Member,~IEEE}

\thanks{This work was supported in part by Jiangxi Province Science and Technology Development Programme under Grant 20242BCC32016; in part by the National Natural Science Foundation of China under Grant 61701197; in part by Basic Research Program of Jiangsu under Grant BK20252084; in part by the National Key Research and Development Program of China under Grant 2021YFA1000500(4) and in part by the 111 Project under Grant B23008. (Jingbo Zhang and Maoxin Ji contributed equally to this work.) (Corresponding author: Qiong Wu.)
	
	Jingbo Zhang, Maoxin Ji and Qiong Wu are with the School of Internet of Things Engineering, Jiangnan University, Wuxi 214122, China, and also with the School of Information Engineering, Jiangxi Provincial Key Laboratory of Advanced Signal Processing and Intelligent Communications, Nanchang University, Nanchang 330031, China (e-mail: jingbozhang@stu.jiangnan.edu.cn; maoxinji@stu.jiangnan.edu.cn; qiongwu@jiangnan.edu.cn).
	
	Pingyi Fan is with the Department of Electronic Engineering, State Key Laboratory of Space Network and Communications, and the Beijing National Research Center for Information Science and Technology, Tsinghua University, Beijing 100084, China (e-mail: fpy@tsinghua.edu.cn).
	
	Kezhi Wang is with the Department of Computer Science, Brunel University, London, Middlesex UB8 3PH, U.K (e-mail: Kezhi.Wang@brunel.ac.uk).
	
	Wen Chen is with the Department of Electronic Engineering, Shanghai Jiao
	Tong University, Shanghai 200240, China (e-mail: wenchen@sjtu.edu.cn).}}

\markboth{Journal of \LaTeX\ Class Files,~Vol.~14, No.~8, October~2025}%
{Shell \MakeLowercase{\textit{et al.}}: A Sample Article Using IEEEtran.cls for IEEE Journals}

\maketitle

\begin{abstract}
Semantic Communication (SC) combined with Vehicular edge computing (VEC) provides an efficient edge task processing paradigm for Internet of Vehicles (IoV). Focusing on highway scenarios, this paper proposes a Tripartite Cooperative Semantic Communication (TCSC) framework, which enables Vehicle Users (VUs) to perform semantic task offloading via Vehicle-to-Infrastructure (V2I) and Vehicle-to-Vehicle (V2V) communications. Considering task latency and the number of semantic symbols, the framework constructs a Mixed-Integer Nonlinear Programming (MINLP) problem, which is transformed into two subproblems. First, we innovatively propose a multi-agent proximal policy optimization task offloading optimization method based on parametric distribution noise (MAPPO-PDN)  to solve the optimization problem of the number of semantic symbols; second, linear programming (LP) is used to solve offloading ratio. Simulations show that performance of this scheme is superior to that of other algorithms.
\end{abstract}

\begin{IEEEkeywords}
Semantic communication, vehicle edge computing, task offloading, multi-agent reinforcement learning.
\end{IEEEkeywords}
\vspace{-5pt}
\section{Introduction}
\IEEEPARstart{I}{n} 
the rapidly moving Internet of Vehicles (IoV), a large number of computation intensive and delay sensitive tasks (such as autonomous driving and vehicle road coordination) put forward strict requirements for low latency and high reliability \cite{ref1,ref2,ref3} , which is difficult to be met solely by on-board terminals \cite{ref4} \cite{ref5} \cite{ref6} . Task offloading and resource allocation can alleviate bottleneck by transferring part of the computation to Road Side Units (RSUs), base stations or neighboring vehicles \cite{ref7} \cite{ref8}.

Traditional bit-level communication\cite{ref9}\cite{ref10} \cite{ref11}\cite{ref12}\cite{ref13} has redundancy and is difficult to meet the requirements of ultra-low latency and low power consumption. As a new paradigm \cite{ref14}, semantic communication (SC) focuses on the semantic validity of information and task relevance, and can significantly reduce the load while ensuring accuracy\cite{ref15}\cite{ref16}. Vehicle edge computing (VEC) has become a key technology for IoV. Through collaborative offloading between vehicles and edge servers, IoV achieves flexible semantic offloading. With SC, VEC enhances communication and computing efficiency as well as system robustness and security \cite{ref17}, offering a new paradigm to meet IoV’s low-latency and high-reliability demands.


Extensive studies have explored SC in Mobile Edge Computing (MEC). For instance, Zheng \textit{et al.} \cite{ref18} proposed a semantics-aware offloading scheme for deep neural network inference, which reduces energy consumption and latency through task feature compression and computation-aware scheduling. Zheng \textit{et al.} \cite{ref17} further introduced a joint optimization approach for computation offloading and semantic compression, improving task completion efficiency. For IoV scenarios, the integration of SC with Vehicle-to-Everything (V2X) collaboration has become a trend. Su \textit{et al.} \cite{ref16} developed a Device-to-Device (D2D) based dynamic resource to enhance spectrum efficiency, while Tang \textit{et al.} \cite{ref7} designed a semantic QoS-driven offloading algorithm balancing similarity and resources. MARL has also been applied to VEC. Zeng \textit{et al.} \cite{ref19} proposed Deep Reinforcement Learning (DRL) based offloading approach for latency-sensitive tasks with dependencies, effectively reducing semantic latency. However, most existing works only consider a single Vehicle-to-Infrastructure (V2I) link and fail to fully leverage advantages brought by Vehicle-to-Vehicle (V2V) collaboration.

To address the aforementioned issues, the main contributions of our work are as follows\footnote{Source code can be found at: \url{https://github.com/qiongwu86/Semantic-Aware-Cooperative-Communication-and-Computation-Framework-in-Vehicular-Networks.git}}:
\begin{itemize}
\item{We propose a flexible tripartite collaborative semantic communication (TCSC) framework integrating V2I/V2V links, enabling dynamic task offloading among local vehicles, RSUs and service vehicles (SVs) for efficient semantic transmission.}
\item{We innovatively propose multi-agent proximal policy optimization based on parametric distribution noise (MAPPO-PDN) with linear programming for semantic communication-driven task offloading, achieving low latency and outperforming traditional MARL.}
\item{
	Simulations show that the proposed MAPPO-PDN outperforms traditional VEC schemes in system delay under different real-world environments.}
\end{itemize}
\vspace{-5pt}
\section{System Model}
\subsection{System Scenario}
The system scenario is illustrated in Fig. 1, where features a two-lane highway. A RSU, connected to a MEC server and covering all lane vehicles, is deployed roadside. Vehicles are categorized into two types: vehicle users (VUs) and SVs. VUs is denoted as \(\mathcal{I} = \{1, 2, \dots, I\}\) and edge node as \(\mathcal{J} = \{0, 1, 2, \dots, J\}\), where 0 denotes the RSU, and $1 \sim J$ denote SVs. All vehicles and RSU are equipped with deep learning-enabled Semantic Communication (DeepSC) systems \cite{ref20}. Time splits into slots \(\mathcal{T} = \{1, 2, \dots, T\}\), each with a duration of $\Delta t$. In slot \(t\), VU task arrivals follow a Poisson distribution, and text-based semantic tasks have size \(y_i\), with VUs as sources. Vehicle speeds follow a truncated Gaussian distribution, and vehicles are uniformly distributed along the road. Each VU preferentially selects the nearest SV as its offloading node. A quasi-static channel is considered, where channel conditions remain constant within a slot but may vary across adjacent slots.

In V2X scenarios, we have discussed the V2I communication model and the V2V communication model. When semantic offloading processing is performed on vehicle tasks, the Signal-to-Interference-plus-Noise Ratio (SINR) between VUs and RSU (or other SVs) can be expressed as
\begin{equation}
	\gamma_{i,j} = \frac{p_{i,j} \left| g_{i,j} \right|}{I_0 + I_{i,j}},
\end{equation}
where \(p_{i,j}\) is transmission power from the $i$-th VU to the $j$-th edge node. \(g_{i,j}\) is the channel gains from the $i$-th TV to the$j$-th edge node, expressed as \(g_{i,j} = \frac{h_0}{d_{i,0}^2}\),  where \(h_0\) is channel gain at distance \(d_0 = 1 \text{m}\), \(d_{i,j}\) is the distance between the $i$-th VU and the $j$-th edge node, expressed as $d_{i,j} = \sqrt{(x_j - x_i)^2 + (y_j - y_i)^2}$. \((x_i, y_i)\) are coordinates of the VU, and \((x_j, y_j)\) are coordinates of the edge node. \(I_0\) represents the noise power. \(I_{i,j} = \sum_{x \neq i} \alpha_{i,j} p_{x,j} \vert g_{x,j} \vert + \sum_{x \neq i} \alpha_{i,j} p_{x,j} \vert g_{x,j} \vert\) is interference from the $i$-th VU to the $j$-th edge node. \(\alpha_{i,j}\) indicates binary spectrum allocation. \(\alpha_{i,j} = 1\) means transmission from the $i$-th VU to the $j$-th edge node uses the same channel.
 \vspace{-10pt}
\subsection{Semantic Communication Model}
We use DeepSC as the text semantic communication model \cite{ref20}. VUs encode text into semantic symbols via DeepSC and send them to the RSU and SVs for semantic decoder \cite{ref18}. 
Unlike traditional data rate, semantic rate, measured in ‘suts/s’, depends on semantic units and information content to reflect semantic information transmitted per unit time\cite{ref19}.

\begin{figure}[t]
	\centering
	\includegraphics[width=\linewidth]{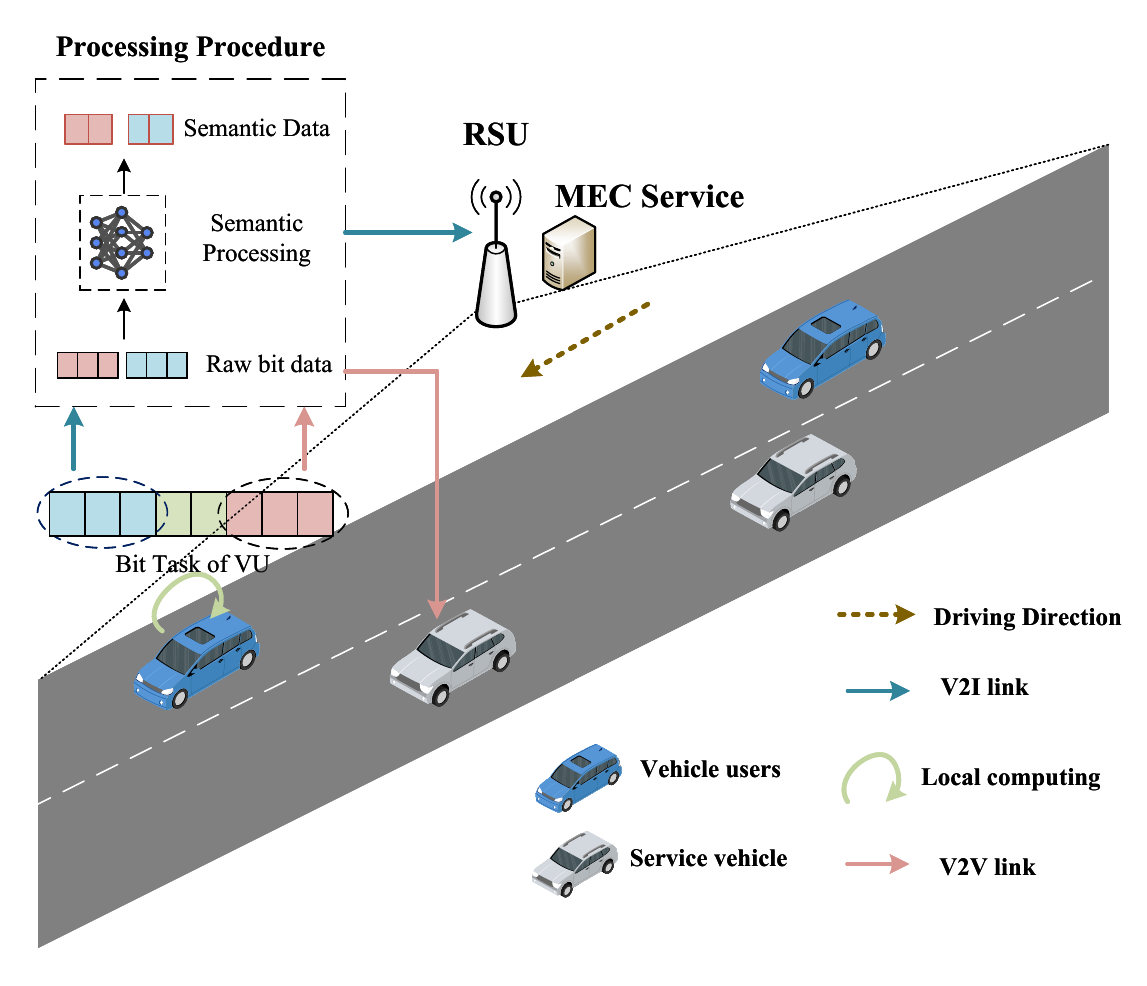} 
	\caption{System overview.}
	\label{fig:fig1}
	\vspace{-0.5cm}
\end{figure}

\begin{equation}
	R_{i,j} = \frac{B S_{i,j}}{L_{i,j} k_{i,j}} \delta_{i,j},
\end{equation}
  where \(B\) is channel bandwidth, \(S_{i,j}\) is average semantic information of a sentence, \(L_{i,j}\) is average length of a sentence, and \(k_{i,j}\) is semantic symbols per word transmissible in a time slot. \(\delta_{i,j}\) is semantic similarity of VU at time slot \(t\), relates to semantic symbols and SINR as \(\delta_{i,j} = f(k_{i,j}, \gamma_{i,j})\), with its mapping relying on DeepSC's neural structure and channel conditions \cite{ref23}. Running a pre-trained DeepSC on an AWGN channel to get the mapping between \(\delta_{i,j}\) and \((k_{i,j}, \gamma_{i,j})\). In \cite{ref23}, a table is formed where the columns are \(k_{i,j}\) and the rows are SINR. The SINR interval is (-10, 20) dB. Since semantic similarity is significantly correlated with the value of \(k_{i,j}\), optimizing \(k_{i,j}\) within this interval can further improve the semantic transmission rate. 
\vspace{-6pt}
\subsection{Computation Model}
Based on the task offloading framework of TCSC, we analyze the computing model of the IoV system.
\subsubsection{Local computing}
 We assume that the local execution ratio of VU in time slot $t$ is \(\rho_i^{\text{local}}\). Therefore, the local computing delay required by VU is calculated as
	\begin{equation}
		T_i^{\text{loc}} = \frac{\rho_i^{\text{local}} y_i C}{f_i^{\text{loc}}},
	\end{equation}
where \(C\) denotes the number of CPU cycles required to compute one bit, \(f_i^{\text{loc}}\) represents computing capability of VU.
\subsubsection{Processed on the RSU}
 We denote offloading ratio of tasks to RSU as \(\rho_i^{\text{RSU}}\). Therefore, transmission delay for VU to offload tasks to the RSU in time slot t can be expressed as
	\begin{equation}
		T_{i,j}^{\text{Tr,RSU}} = \frac{\rho_i^{\text{RSU}} \, d_i \, S_{i,j}}{R_{i,j}} = \frac{\rho_i^{\text{RSU}} \, d_i \, k_{i,j} \, L_{i,j}}{B \, \delta_{i,j}},
	\end{equation}
	where \(d_i\) is the number of semantic task queues of VU in time slot t, expressed as \(d_i = y_i / H\). Here, \(H\) is conversion factor that converts the data volume \(y_i\) into the number of sentences. The computation delay for VU to offload to the RSU can be calculated as
	\begin{equation}
		T_{i,j}^{\text{RSU}} = \frac{\rho_{i}^{\text{RSU}} \, y_i \, C}{f_{i,j}^{\text{RSU}} / N},
	\end{equation}
	Considering that the RSU needs to allocate computing capability to all connected VUs, we set the total computing capability of the RSU as \(f_{i,j}^{\text{RSU}}\) and have it shared by \(N\) VUs.
	
After processing, the RSU returns the results to the corresponding VU. Since the result size is negligible, downlink transmission delay is ignored.
\subsubsection{Processed on the SV}
We denote the offloading ratio of tasks to $j$-th SV as \(\rho_i^{\text{SV}}\). The transmission delay for VU to offload tasks to $j$-th SV in time slot $t$ can be expressed as
	\begin{equation}
		T_{i,j}^{\text{Tr,SV}} = \frac{\rho_{i}^{\text{SV}} \, d_i \, S_{i,j}}{R_{i,j}} = \frac{\rho_{i}^{\text{SV}} \, d_i \, k_{i,j} \, L_{i,j}}{B \, \delta_{i,j}}.
	\end{equation}
	
VU offloads a fraction \(\rho_i^{\text{SV}}\) of its task to the nearest $j$-th SV, the available computing resource depends on $j$-th SV capacity and the number of connected tasks. Let the total capacity of $j$-th SV be \(f_{i,j}^{\text{SV}}\), with $N$ VUs and $J$ SVs. If $M$ VU pick $j$-th SV as the computing node in the same time slot, $j$-th SV evenly allocates resources to them. $M$ has a threshold \(0 < M < N/J\); if exceeded, VUs offload to the second nearest SV. The computation delay for VU offloading to SV is
	\begin{equation}
		T_{i,j}^{\text{SV}} = \frac{\rho_{i}^{\text{SV}} \, y_i \, C}{f_{i,j}^{\text{SV}} / M}.
	\end{equation}
	
	Similar to the aforementioned RSU offloading section, the downlink delay for returning these results is not considered. Finally, based on the task offloading decision for VU, the total delay of task $y_i$ is given by the following formula
	\begin{equation}
		T_{i}^{\text{tot}} = \max\left\{ T_{i}^{\text{loc}}, \, T_{i,j}^{\text{tr,RSU}} + T_{i,j}^{\text{RSU}}, \, T_{i,j}^{\text{tr,SV}} + T_{i,j}^{\text{SV}} \right\}.
	\end{equation}
\subsection{Problem Formulation}
Considering that the offloading ratio \(\rho\) and the number of semantic symbols \(k\) under the TCSC framework are adjustable, we minimize the total delay of all vehicles by jointly adjusting them. The optimization problem can be formulated as
\begin{subequations}
	\begin{align}
		&\textbf{P1: } \min_{\rho, k} \sum_{i \in \mathcal{I}} T_{i}^{\text{tot}} \label{eq:objective} \\
		&\text{s.t.} \quad \rho_{i}^{\text{loc}}, \rho_{i}^{\text{RSU}}, \rho_{i}^{\text{SV}} \in [0,1], \quad \forall i \label{eq:C1} \\
		&\qquad \rho_{i}^{\text{loc}} + \rho_{i}^{\text{RSU}} + \rho_{i}^{\text{SV}} = 1, \quad \forall i \label{eq:C2} \\
		&\qquad \delta_{i,j} \geq \delta_{\text{th}}, \quad \forall i,j \label{eq:C3} \\
		&\qquad  k_{i,j} \in \{1, \dots, k_{i,j}^{\text{max}}\} \label{eq:C4} \\
		&\qquad T_{i}^{\text{loc}} \leq D_{i}^{\text{max}}, \quad \forall i \label{eq:C5} \\
		&\qquad T_{i,j}^{\text{tr,RSU}} + T_{i,j}^{\text{RSU}} \leq D_{i}^{\text{max}}, \quad \forall i,j \label{eq:C6} \\
		&\qquad T_{i,j}^{\text{tr,SV}} + T_{i,j}^{\text{SV}} \leq D_{i}^{\text{max}}, \quad \forall i,j \label{eq:C7}
	\end{align}
\end{subequations}

In P1, constraint (9b) denotes task allocation ratios; (9c) ensures complete distribution across local, RSU, and SV execution; (9d) guarantees semantic similarity in V2I/V2V links exceeds a threshold; (9e) limits the average number of semantic symbols per word. Constraints (9f), (9g), and (9h) ensure each task portion meets delay constraints.

Problem P1 is hard to solve as the objective function couples the continuous offloading ratio \(\rho\) and discrete semantic symbol count \(k_{i,j}\). Using only Multi-Agent Proximal Policy Optimization (MAPPO) needs extra design to disperse continuous variables, raising complexity and lowering accuracy. Therefore, following \cite{ref24}, we decompose the problem. First, based on the parameter uncertainty regularization concept in \cite{ref25}, we propose a MAPPO-PDN for multi-agent to optimize discrete semantic symbol counts. Introducing parametric distribution noise (PDN) boosts generalization and convergence stability. Second, linear programming (LP) is used to optimize continuous offloading ratio, thereby reducing the overall complexity.

\vspace{-5 pt}
\section{Problem Formulation and Solution}
\subsection{Semantic Symbol Optimization Based on MAPPO-PDN}
To select the optimal number of semantic symbols \(k_{i,j}\), we use the MAPPO-PDN algorithm to optimize the discrete variable $k_{i,j}$ in (9a). This algorithm is a modified MAPPO. Therefore, the constraints only related to the value of \(k_{i,j}\), and are kept by fixing \(\rho\), we transform original problem into:

\vspace{-5 pt}
\begin{equation} \label{eq:your_main_label}
	\begin{aligned}
		\textbf{P2: } & \min_{k} \sum_{i \in \mathcal{I}} T_{i}^{\text{tot}} \\
		\text{s.t. } & \text{(9d)}, \text{(9e)}, \text{(9g)}, \text{(9h)}
	\end{aligned}
\end{equation}

Training the number of semantic symbols using the MAPPO-PDN algorithm, all VUs act as agents. Through multi-agent interaction and adjusting the number of semantic symbols according to policies, it can provide the solution to the optimization problem P2. Generally, it consists of state space \(s\), action space \(a\), and reward function \(r\).

\subsubsection{State space}
State space can be expressed as \(s_t = \{ \gamma_{i,j}, \hat{d}_{i,j}, y_i, \mathcal{F}_{\gamma} \}\), where \(\gamma_{i,j}\) represents the SINR, \(\hat{d}_{i,j}\) is the normalization of the distance \(d_{i,j}\), \(y_i\) is the task size, and $\mathcal{F}_{\gamma}$ is a vector composed of semantic similarity values obtained from the mapping function \(\delta_{i,j} = f(k_{i,j}, \gamma_{i,j})\). Different dimensions represent the semantic similarity corresponding to different values of $k$ under the current SINR condition. Each agent makes decisions independently based on state information.
\subsubsection{Action space}
All agents select the optimal action based on the currently observed state \(s_t\). While SINR is in the range of (-10, 20) dB, this action can be expressed as \(a_t = \{1, \dots, k_{i,j}^{\text{max}}\}\), where \(k_{i,j}^{\text{max}}\) represents the maximum average number of semantic symbols used for each word. If SINR is lower than the minimum threshold we trained, \(k_{i,j}\) takes \(k_{i,j}^{\max}\); if SINR is higher than the maximum threshold we trained, \(k_{i,j}\) takes 1. This is because when the SINR is good, a relatively small $k$ is sufficient to achieve a high semantic similarity. However, when the SINR is poor, a larger k value is needed to ensure the semantic similarity.
\subsubsection{Reward function}
The reward \(r_t\) refers to the immediate reward obtained when action \(a_t\) is executed under state \(s_t\), and the reward function is designed as \(r_t = T_{\text{min}} / T_{i}\), where $T_{\text{min}}$ is minimum delay obtained by iterating over all $k_{i,j}$ using Equation (8). $T_{i}$ is delay calculated based on the action.

Different from the point estimation method for policy parameters in traditional MAPPO, our method models the parameters \(\theta_k\) of the Actor as a Gaussian distribution \(q(\theta_k) = \mathcal{N}(\mu_k, \sigma_k^2)\), and samples decisions for each agent \(n\) at each time \(t\) via reparameterization $\theta_k^{(n)} = \mu_k + \sigma_k \xi_k^{(n)}$. Here, \(k\) denotes the parameter index, \(n\) denotes the agent index. All agents share \((\mu_k, \sigma_k)\) but sample \(\xi_k^{(n)}\) independently, its effect will be truncated in the following processing. This design enables policy to directly learn parameter groups, thereby explicitly modeling uncertainty and maintaining strong robustness in scenarios with environmental changes and out-of-distribution cases. This parallel posterior sampling mechanism not only improves exploration diversity but reduces overfitting risk in non-stationary multi-agent environments. Drawing on idea of PDN, we introduce a PDN upper bound constraint into the MAPPO loss \cite{ref25}:
\vspace{-5 pt}
\begin{equation}
\mathcal{L}_{SNR} = \mathbb{E}\left[\left( \max\left( \Omega_k - \Omega_{\text{max}}, 0 \right) \right)^2\right],
\end{equation}
We define PDN of a specific parameter as \(\Omega_k = \frac{|\mu_k|}{\sigma_k}\). Where $\Omega_{\text{max}}$ is upper limit for PDN. Complex and unstable environments require a larger \(\Omega_{\text{max}}\). It limits the upper bound of parameter uncertainty, ensuring exploration diversity and improving generalization ability. The clipped policy loss function of MAPPO is:

\vspace{-6 pt}
\begin{equation}
	\mathcal{L}_{\pi} = -\mathbb{E} \left[ \min \left( \rho_t \hat{A}_t, \text{clip} \left( \rho_t, 1 - \varepsilon, 1 + \varepsilon \right) \hat{A}_t \right) \right],
\end{equation}
where \( \rho_{t}^{(n)} = \frac{\pi_{\theta} \left( a_{t}^{(n)} \big| \boldsymbol{s}_{t}^{(n)} \right)}{\pi_{\theta_{\text{old}}} \left( a_{t}^{(n)} \big| \boldsymbol{s}_{t}^{(n)} \right)} \) is importance sampling ratio, \( \pi_{\theta} \) is actor policy, and \( \pi_{\theta_{\text{old}}} \) is old policy for sampling data. \(\hat{A}_t\) is advantage function. \(\text{clip}(\rho_{t}, 1 - \varepsilon, 1 + \varepsilon)\) clips \(\rho_t^{(n)}\) to \([1-\epsilon, 1+\epsilon]\). The value function loss is the mean squared error between the predicted state value and the actual return, expressed as
\begin{equation}
	\mathcal{L}_{V} = \frac{1}{2} \mathbb{E} \left[ \left( V_{\psi}(\boldsymbol{s}_t) - \hat{V}_t \right)^2 \right],
\end{equation}
where \(\hat{V}_t = \hat{A}_t + V_{\psi}(\boldsymbol{s}_t)\), and \(\hat{V}_t\) is the regression target value constructed by the critic during training. To improve the exploration ability of MAPPO, we define the regularization function as \(\mathcal{L}_{\mathcal{H}} = -\mathbb{E} \left[ \mathcal{H} \left( \pi_{\theta} (\cdot \mid \boldsymbol{s}_t) \right) \right]\), where \(\mathcal{H}(\pi_{\theta_{\text{old}}})\) is the strategy entropy. Finally, PDN regularization is added to the loss of MAPPO to form our method, MAPPO-PDN.

\vspace{-5 pt}
\begin{equation}
	\mathcal{L} = \mathcal{L}_{\pi} + c_v \mathcal{L}_{V} + \beta_t \mathcal{L}_{\mathcal{H}} + \omega \mathcal{L}_{SNR},
\end{equation}
\(c_v\), \(\beta_t\) and \(\omega\) are weight of value loss, weight of entropy regularization, and weight of PDN regularization, respectively.
\vspace{-15 pt}
\subsection{Offloading Rate Based on Linear Programming}
To select the optimal offloading rate \(\rho\) with lower complexity and higher accuracy, we use LP to solve the other part.
\begin{subequations} 
	\begin{align}
		\text{P3}: &\quad \min_{\rho} \sum_{i \in \mathcal{I}} T_i^{\text{tot}} \\
		&\quad \text{s.t.}\ \text{(10b)}, \text{(10c)}, \text{(10f)}, \text{(10g)}, \text{(10h)}
	\end{align}
\end{subequations}
Constraints (10f), (10g), and (10h) are all inequalities related to \(\rho\). We set \( A_L = \frac{y_i C}{f_i^{\text{loc}}} \), \( A_R = \frac{d_i k_{i,j} L_{i,j}}{B \delta_{i,j}} + \frac{y_i C}{f_{i,j}^{\text{RSU}} / N} \) and  \( A_S = \frac{d_i k_{i,j} L_{i,j}}{B \delta_{i,j}} + \frac{y_i C}{f_{i,j}^{\text{SV}} / M} \), they can be simplified as
\begin{equation}
	\rho_i^{\text{local}} \left( \frac{y_i C}{f_i^{\text{loc}}} \right) = \rho_i^{\text{local}} A_L \leq D_i^{\text{max}}
\end{equation}
\begin{equation}
	\rho_i^{\text{RSU}} \left( \frac{d_i k_{i,j} L_{i,j}}{B \delta_{i,j}} + \frac{y_i C}{f_{i,j}^{\text{RSU}} / N} \right) = \rho_i^{\text{RSU}} A_R \leq D_i^{\text{max}}
\end{equation}
\begin{equation}
	\rho_i^{\text{SV}} \left( \frac{d_i k_{i,j} L_{i,j}}{B \delta_{i,j}} + \frac{y_i C}{f_{i,j}^{\text{SV}} / M} \right) = \rho_i^{\text{SV}} A_S \leq D_i^{\text{max}}
\end{equation}

To find the optimal solution, all constraints should be tight. Therefore, we have \(\rho_i^{\text{local}} A_L = D_i^{\text{max}*}\), \(\rho_i^{\text{RSU}} A_R = D_i^{\text{max}*}\), and \(\rho_i^{\text{SV}} A_S = D_i^{\text{max}*}\). According to constraint C2, we obtain
\begin{equation}
	D_i^{\text{max}*} \left( \frac{1}{A_L} + \frac{1}{A_R} + \frac{1}{A_S} \right) = 1
\end{equation}

Since the \( k \) value has been given by the previous MAPPO-SNR algorithm, \( \frac{1}{A_L} + \frac{1}{A_R} + \frac{1}{A_S} \) is a known value, and \(D_i^{\text{max}*}\) can be obtained. Therefore, according to (17), (18), and (19), all offloading rates \( \rho \) can be derived: \( \rho_i^{\text{local}*} = \frac{D_i^{\text{max}*}}{A_L} \), \( \rho_i^{\text{RSU}*} = \frac{D_i^{\text{max}*}}{A_R} \), and \( \rho_i^{\text{SV}*} = \frac{D_i^{\text{max}*}}{A_S} \).

This method combines MAPPO-PDN with LP to design a TCSC-based MAPPO-PDN-LP offloading strategy, which can balance semantic symbol optimization and task allocation, and improve the overall task processing efficiency of the system.
\begin{figure}[t]
	\centering
	\includegraphics[width=0.85\linewidth]{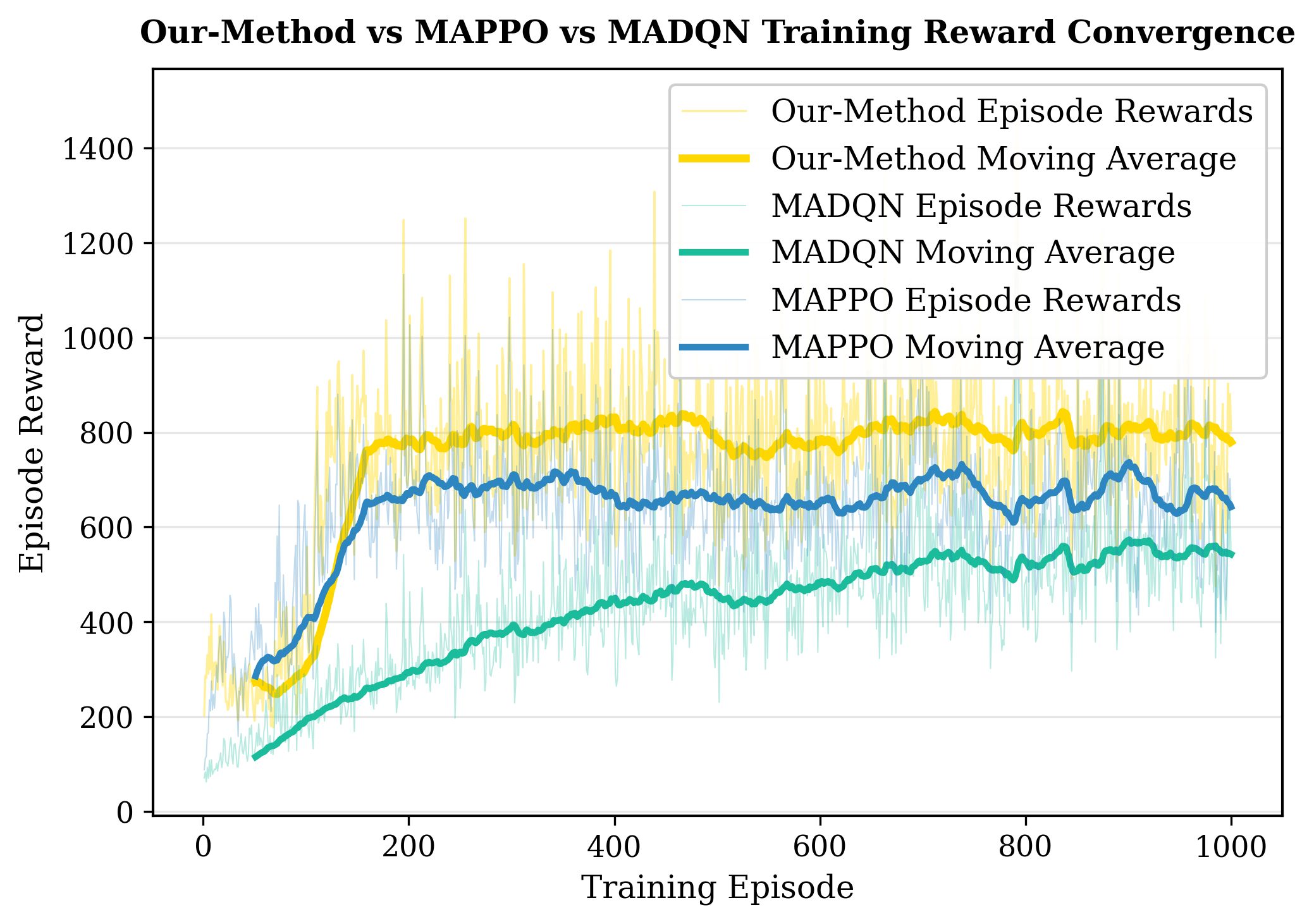} 
	\caption{Convergence performance comparisons.}
	\label{fig:fig}
	\vspace{-0.4cm}
\end{figure}
\section{Simulation Results and Analysis}
This section evaluates the performance of proposed TCSC method through simulation, which includes one RSU and vehicles distributed along a 400m road. When tasks are allocated to VUs, each time slot generates a task volume of 0.8 Mbit, which is generated by a Poisson distribution, and the required computing frequency \textit{C} is 1000 cycles. We set the number of VUs are 20, the number of SVs is fixed at \textit{J} = 5. Vehicle speeds are initialized by uniform sampling within 50 to 80km/h and assumed constant thereafter. Vehicle speed is for theoretical modeling and training settings; in actual simulations, vehicle speed is constant, and inter-vehicle spacing is related to the number of vehicles. Noise power is $-114$dBm, V2V/V2I transmit powers are 15dBm and 23dBm, respectively, and the bandwidth $B$ is 540kHz. RSU, VUs and SVs’ computing capacities \(f_{i,j}^{\text{RSU}}\), \(f_i^{\text{loc}}\), \(f_{i,j}^{\text{SV}}\) are 6GHz, 1GHz and 3GHz. Additionally, the average length \(L_{i,j}\) of each sentence is set to 20, conversion factor \(H\) is 1200 bit, and semantic similarity threshold $\delta_{th}$ is 0.9. $\Omega_{\text{max}}$ is set to 20. Experiments are conducted using Python 3.8 and PyTorch.

The proposed MAPPO-PDN+LP method is compared with the following methods: 1) MAPPO-PDN + lvy: Collaborate MAPPO-PDN with lvy \cite{ref26} for semantic offloading. 2) MAPPO + LP: Collaborate MAPPO with LP for semantic offloading. 3) MADQN + LP: Collaborate MADQN with LP for semantic offloading. 4) Linear K-Selection: Linear selection of $k$ values, combined with LP for semantic offloading. 5) Traditional + LP: Apply the LP method to traditional communication offloading. 6) Traditional + lvy: Apply the lvy method to traditional communication offloading. 7) Traditional (No LP): Traditional communication offloading is performed with a fixed offloading ratio. 

\begin{figure}[t]
	\centering
	\includegraphics[width=0.85\linewidth]{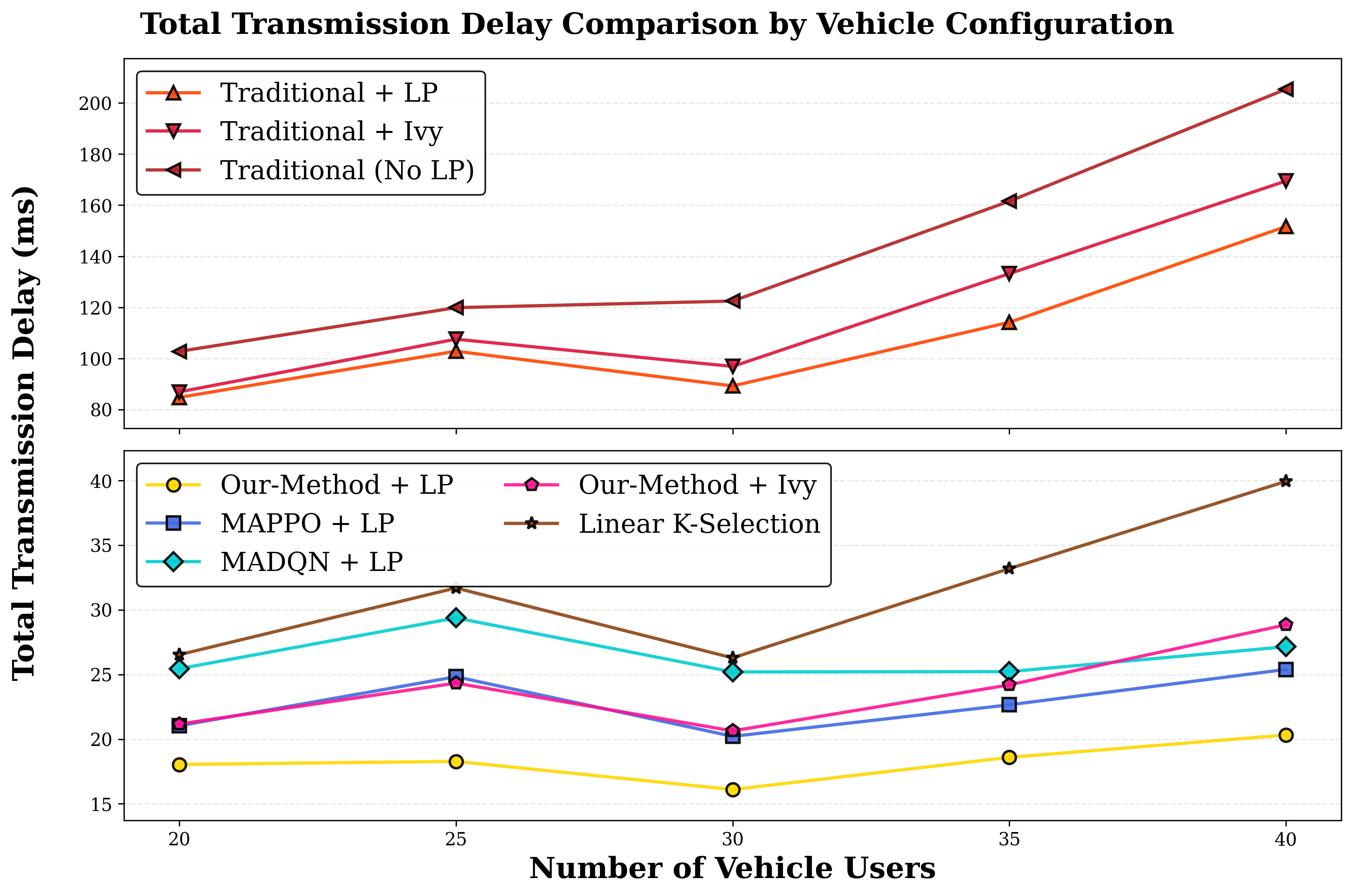} 
	\caption{Average transmission delay versus vehicular number.}
	\label{fig:fig}
	\vspace{-0.2cm}
\end{figure}

Fig. 2 shows the reward convergence performance of proposed MAPPO-PDN method and other benchmark methods. Traditional MARL is prone to overfitting or strategy instability due to point estimation. However, MAPPO-PDN enhances adaptability in non-stationary environments by means of parameter uncertainty modeling and PDN regularization, continuously obtaining high rewards during training, and its rewards are significantly higher than those of traditional MARL.
\begin{figure}[t]
	\centering
	\includegraphics[width=0.85\linewidth]{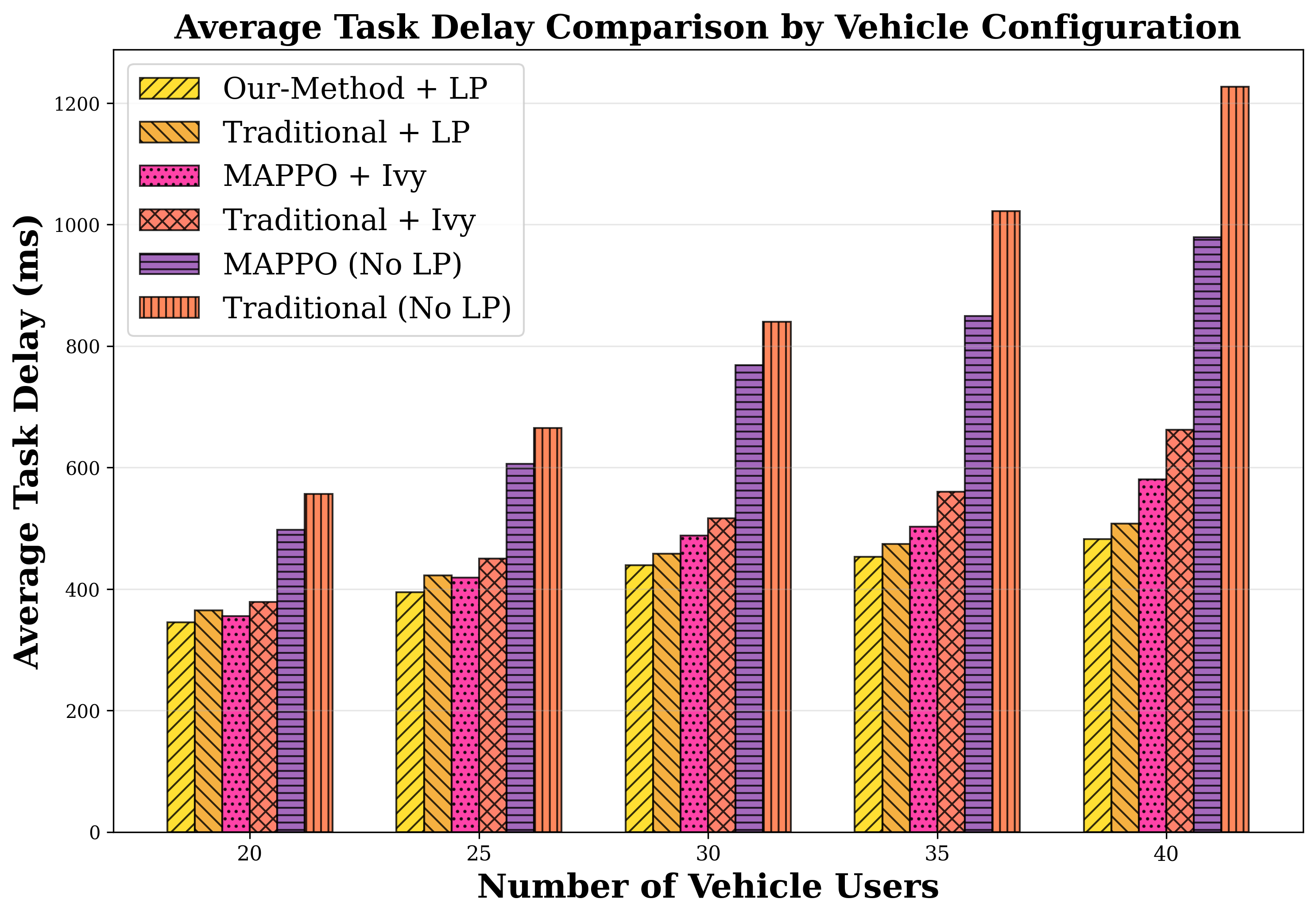} 
	\caption{Average task delay versus vehicular number.}
	\label{fig:fig}
	\vspace{-0.4cm}
\end{figure}

Fig. 3 plots total transmission delay under different VU numbers. The vehicle speed remains constant. As vehicle rises, overall delay trends upward due to limited resource blocks. The transmission delay exhibits a concave downward trend when vehicle number is 25 to 35. This is because more vehicles mean closer VU and SV distance, lower transmission loss, higher rate, reducing delay. But when vehicle number hits a certain level (delay peaks at 40 vehicles as VUs reuse resource blocks more), delay rises. Compared with other methods, our method has lower latency.

Fig. 4 presents a bar chart of the average total task delay under different numbers of VUs. The vehicle keeps moving at a constant speed. It can be observed that as the number of vehicles increases, the average task delay shows an upward trend. This is because, in addition to the relationship between the number of vehicles and transmission delay discussed in Fig. 3 above, there is also a component of computational delay. As the number of vehicles increases, the computing resources of edge nodes are allocated on average, which leads to an increase in computational delay and thus a higher total delay. It can also be seen that the overall total delay mainly depends on the computational delay. In conclusion, it is evident that the proposed method demonstrates better performance and lower delay compared with other methods.
\vspace{-5pt}
\section{Conclusion}
This paper discussed the task offloading problem of collaborative semantic offloading using joint V2I/V2V in VEC systems within high-speed moving IoV. Based on the TCSC framework the MAPPO-PDN algorithm, combined with LP, was proposed to minimize the total delay by jointly optimizing the number of semantic symbols and offloading rate. Simulation results demonstrate the effectiveness of our proposed algorithm. In the future, we will explore multimodal semantic communication in IoV scenarios.

\newpage

\vfill


\begin{thebibliography}{1}
\bibliographystyle{IEEEtran}

\bibitem{ref1}
Mao W, Xiong K, Lu Y, et al. Energy consumption minimization in secure multi-antenna UAV-assisted MEC networks with channel uncertainty[J]. IEEE Transactions on Wireless Communications, 2023, 22(11): 7185-7200.

\bibitem{ref2}
Dong Y, Chen Z, Liu S, et al. Age-upon-decisions minimizing scheduling in Internet of Things: To be random or to be deterministic?[J]. IEEE Internet of Things Journal, 2019, 7(2): 1081-1097.

\bibitem{ref3}
Di X, Xiong K, Fan P, et al. Optimal resource allocation in wireless powered communication networks with user cooperation[J]. IEEE Transactions on Wireless Communications, 2017, 16(12): 7936-7949.

\bibitem{ref4}
Q. Wu and J. Zheng, "Performance modeling of IEEE 802.11 DCF based fair channel access for vehicular-to-roadside communication in a non-saturated state," 2014 IEEE International Conference on Communications (ICC), Sydney, NSW, Australia, 2014, pp. 2575-2580, doi: 10.1109/ICC.2014.6883711.

\bibitem{ref5}
Shao Z, Wu Q, Fan P, et al. Semantic-aware resource allocation based on deep reinforcement learning for 5g-v2x hetnets[J]. IEEE Communications Letters, 2024.

\bibitem{ref6}
Qi K, Wu Q, Fan P, et al. Reconfigurable-intelligent-surface-aided vehicular edge computing: Joint phase-shift optimization and multiuser power allocation[J]. IEEE Internet of Things Journal, 2024, 12(1): 764-777.

\bibitem{ref7}
C. Tang and N. Zhang, ``Semantic QoS-aware Resource Allocation for IoV Computation Offloading using Layer Enhanced Whale Optimization Algorithm,'' in \textit{Proc. 2024 10th Int. Conf. Comput. Commun. (ICCC)}, Chengdu, China, 2024, pp. 1647--1651. DOI: 10.1109/ICCC62609.2024.10941942.

\bibitem{ref8}
M. Zheng, H. Yang, S. Liu, K. Lin, L. Xiao, and Z. Han, ``Reliable Semantic Communication With QoE-Driven Resource Scheduling for UAV-Assisted MEC,'' \textit{IEEE Trans. Veh. Technol.}, vol. 74, no. 7, pp. 11484--11489, Jul. 2025. DOI: 10.1109/TVT.2025.3542775.

\bibitem{ref9}
Li T, Fan P, Chen Z, et al. Optimum transmission policies for energy harvesting sensor networks powered by a mobile control center[J]. IEEE Transactions on Wireless Communications, 2016, 15(9): 6132-6145.

\bibitem{ref10}
Yang Y, Fan P. Doppler frequency offset estimation and diversity reception scheme of high-speed railway with multiple antennas on separated carriage[J]. Journal of Modern Transportation, 2012, 20(4): 227-233.

\bibitem{ref11}
Zhou H, Fan P, Li J. Global proportional fair scheduling for networks with multiple base stations[J]. IEEE Transactions on Vehicular Technology, 2011, 60(4): 1867-1879.

\bibitem{ref12}
Yao Z, Jiang J, Fan P, et al. A neighbor-table-based multipath routing in ad hoc networks[C]//The 57th IEEE Semiannual Vehicular Technology Conference, 2003. VTC 2003-Spring. IEEE, 2003, 3: 1739-1743.

\bibitem{ref13}
Fan P, Feng C, Wang Y, et al. Investigation of the time-offset-based QoS support with optical burst switching in WDM networks[C]//2002 IEEE International Conference on Communications. Conference Proceedings. ICC 2002 (Cat. No. 02CH37333). IEEE, 2002, 5: 2682-2686.

\bibitem{ref14}
Z. Shao \textit{et al.}, ``Semantic-Aware Spectrum Sharing in Internet of Vehicles Based on Deep Reinforcement Learning,'' \textit{IEEE Internet Things J.}, vol. 11, no. 23, pp. 38521--38536, Dec. 2024.

\bibitem{ref15}
Z. Ji, Z. Qin, X. Tao, and Z. Han, ``Resource Optimization for Semantic-Aware Networks With Task Offloading,'' \textit{IEEE Trans. Wireless Commun.}, vol. 23, no. 9, pp. 12284--12296, Sept. 2024. 

\bibitem{ref16}
J. Su \textit{et al.}, ``Semantic Communication-Based Dynamic Resource Allocation in D2D Vehicular Networks,'' \textit{IEEE Trans. Veh. Technol.}, vol. 72, no. 8, pp. 10784--10796, Aug. 2023. DOI: 10.1109/TVT.2023.3257770.

\bibitem{ref17}
Y. Zheng, T. Zhang, R. Huang, and Y. Wang, ``Computing Offloading and Semantic Compression for Intelligent Computing Tasks in MEC Systems,'' in \textit{Proc. IEEE Wireless Commun. Netw. Conf. (WCNC)}, Glasgow, U.K., 2023, pp. 1--6. DOI: 10.1109/WCNC55385.2023.10118995.

\bibitem{ref18}
G. Zheng, M. Wen, Z. Ning, and Z. Ding, ``Computation-Aware Offloading for DNN Inference Tasks in Semantic Communication Assisted MEC Systems,'' \textit{IEEE Trans. Wireless Commun.}, vol. 24, no. 4, pp. 2693--2706, Apr. 2025. DOI: 10.1109/TWC.2024.3523517.

\bibitem{ref19}
J. Zeng, Z. Kuang, R. Chen, and A. Liu, ``Delay-Sensitive Dependent Tasks Offloading and Resource Allocation in VEC: A Deep-Reinforcement Learning Approach,'' \textit{IEEE Internet Things J.}, vol. 12, no. 16, pp. 34190--34203, Aug. 15, 2025. DOI: 10.1109/JIOT.2025.3577449.

\bibitem{ref20}
H. Xie, Z. Qin, G. Y. Li, and B.-H. Juang, ``Deep Learning Enabled Semantic Communication Systems,'' \textit{IEEE Trans. Signal Process.}, vol. 69, pp. 2663--2675, 2021. DOI: 10.1109/TSP.2021.3071210.


\bibitem{ref21}
N. Reimers and I. Gurevych, ``Sentence-BERT: Sentence Embeddings using Siamese BERT-Networks,'' in \textit{Proc. Conf. Empir. Methods Nat. Lang. Process. (EMNLP)}, 2019. [Online]. 

\bibitem{ref22}
Z. Ji and Z. Qin, ``Energy-Efficient Task Offloading for Semantic-Aware Networks,'' in \textit{Proc. IEEE Int. Conf. Commun. (ICC)}, Rome, Italy, 2023, pp. 3584--3589. DOI: 10.1109/ICC45041.2023.10279646.

\bibitem{ref23}
L. Yan, Z. Qin, R. Zhang, Y. Li, and G. Y. Li, ``Resource Allocation for Text Semantic Communications,'' \textit{IEEE Wireless Commun. Lett.}, vol. 11, pp. 1394--1398, 2022. [Online]. 

\bibitem{ref24}
N. Lin, H. Tang, L. Zhao, S. Wan, A. Hawbani and M. Guizani, "A PDDQNLP Algorithm for Energy Efficient Computation Offloading in UAV-Assisted MEC," in IEEE Transactions on Wireless Communications, vol. 22, no. 12, pp. 8876-8890, Dec. 2023.

\bibitem{ref25}
P. Moure, L. Cheng, J. Ott, Z. Wang and S. -C. Liu, "Regularized Parameter Uncertainty for Improving Generalization in Reinforcement Learning," 2024 IEEE/CVF Conference on Computer Vision and Pattern Recognition (CVPR), Seattle, WA, USA, 2024, pp. 23805-23814.


\bibitem{ref26}
M. Ghasemi, \textit{et al.}, "Optimization based on the smart behavior of plants with its engineering applications: Ivy algorithm," Knowledge-Based Systems, 2024, ISSN 0950-7051, https://doi.org/10.1016/j.knosys.2024.111850.
\end{thebibliography}
\end{document}